\PassOptionsToPackage{dvipsnames}{xcolor}

\documentclass[sigconf, screen, balance]{acmart}

\AtBeginDocument{%
  }

\setcopyright{acmlicensed}
\copyrightyear{2018}
\acmYear{2018}
\acmDOI{XXXXXXX.XXXXXXX}
\acmConference[Conference acronym 'XX]{Make sure to enter the correct
  conference title from your rights confirmation email}{June 03--05,
  2018}{Woodstock, NY}
\acmISBN{978-1-4503-XXXX-X/2018/06}

\usepackage{hyperref}       
\usepackage{url}            
\usepackage{booktabs}       
\usepackage{amsfonts}       
\usepackage{nicefrac}       
\usepackage{microtype}      
\usepackage{color}

\usepackage{multirow}
\usepackage{wrapfig}
\usepackage{graphicx}
\usepackage{mathtools}
\usepackage{subcaption}
\usepackage{amsmath}
\usepackage{amsthm}
\usepackage{algorithm}
\usepackage{algpseudocode}
\usepackage{colortbl}
\usepackage{adjustbox}
\usepackage{cuted}
\usepackage{multicol}
\usepackage{cleveref}
\usepackage{svg}
\usepackage{enumitem}
\usepackage{makecell}

\begin{document}

\title{WisWheat: A Three-Tiered Vision-Language Dataset for Wheat Management}

\author{Bowen Yuan}
\authornote{Both authors contributed equally to this research.}
\email{bowen.yuan@uq.edu.au}
\affiliation{%
  \institution{The University of Queensland}
  \city{Brisbane}
  \state{Queensland}
  \country{Australia}
}

\author{Selena Song}
\authornotemark[1]
\email{selena.song@uq.edu.au}
\affiliation{%
  \institution{The University of Queensland}
  \city{Brisbane}
  \state{Queensland}
  \country{Australia}
}

\author{Javier Fernandez}
\email{j.fernandez@uq.edu.au}
\affiliation{%
  \institution{The University of Queensland}
  \city{Brisbane}
  \state{Queensland}
  \country{Australia}
}

\author{Yadan Luo}
\email{y.luo@uq.edu.au}
\affiliation{%
  \institution{The University of Queensland}
  \city{Brisbane}
  \state{Queensland}
  \country{Australia}
}

\author{Mahsa Baktashmotlagh}
\email{m.baktashmotlagh@uq.edu.au}
\affiliation{%
  \institution{The University of Queensland}
  \city{Brisbane}
  \state{Queensland}
  \country{Australia}
}

\author{Zijian Wang}
\email{zijian.wang@uq.edu.au}
\affiliation{%
  \institution{The University of Queensland}
  \city{Brisbane}
  \state{Queensland}
  \country{Australia}
}

\renewcommand{\shortauthors}{Yuan et al.}
\newcommand{\eat}[1]{}

\begin{abstract}


Wheat management strategies play a critical role in determining yield. Traditional management decisions often rely on labour-intensive expert inspections, which are expensive, subjective and difficult to scale. Recently, Vision-Language Models (VLMs) have emerged as a promising solution to enable scalable, data-driven management support. However, due to a lack of domain-specific knowledge, directly applying VLMs to wheat management tasks results in poor quantification and reasoning capabilities, ultimately producing vague or even misleading management recommendations. In response, we propose WisWheat, a wheat-specific dataset with a three-layered design to enhance VLM performance on wheat management tasks: (1) a foundational pretraining dataset of 47,871 image–caption pairs for coarsely adapting VLMs to wheat morphology; (2) a quantitative dataset comprising 7,263 VQA-style image–question–answer triplets for quantitative trait measuring tasks; and (3) an Instruction Fine-tuning dataset with 4,888 samples targeting biotic and abiotic stress diagnosis and management plan for different phenological stages. Extensive experimental results demonstrate that fine-tuning open-source VLMs (e.g., Qwen2.5 7B) on our dataset leads to significant performance improvements. Specifically, the Qwen2.5 VL 7B fine-tuned on our wheat instruction dataset achieves accuracy scores of 79.2\% and 84.6\% on wheat stress and growth stage conversation tasks respectively, surpassing even general-purpose commercial models such as GPT-4o by a margin of 11.9\% and 34.6\%. 

\end{abstract}

\ccsdesc[500]{Computing methodologies~Computer vision tasks}

\keywords{Vision Language Dataset, Vision Language Model}


\maketitle

\section{Introduction}
As one of the most widely cultivated staple crops, wheat contributes to nearly 20\% of the global dietary caloric intake and serves as a central component of food systems worldwide~\cite{erenstein2022global}. Improving wheat productivity and resilience requires a comprehensive understanding of the interactions among genotype, environment, and management, which are collectively known as G$\times$E$\times$M~\cite{gxexm}. While advancements in breeding and environmental monitoring have received significant attention~\cite{gxe1, gxe2}, the role of agronomic management remains an under-leveraged but critical lever for optimising crop performance. Management practices, such as wheat phenotyping, monitoring and stress mitigation, are actionable interventions that enable growers to adapt to local environmental conditions and unlock the full genetic potential of wheat cultivars. 

\begin{figure*}[t]
\centering
    \includegraphics[width=0.9\textwidth]{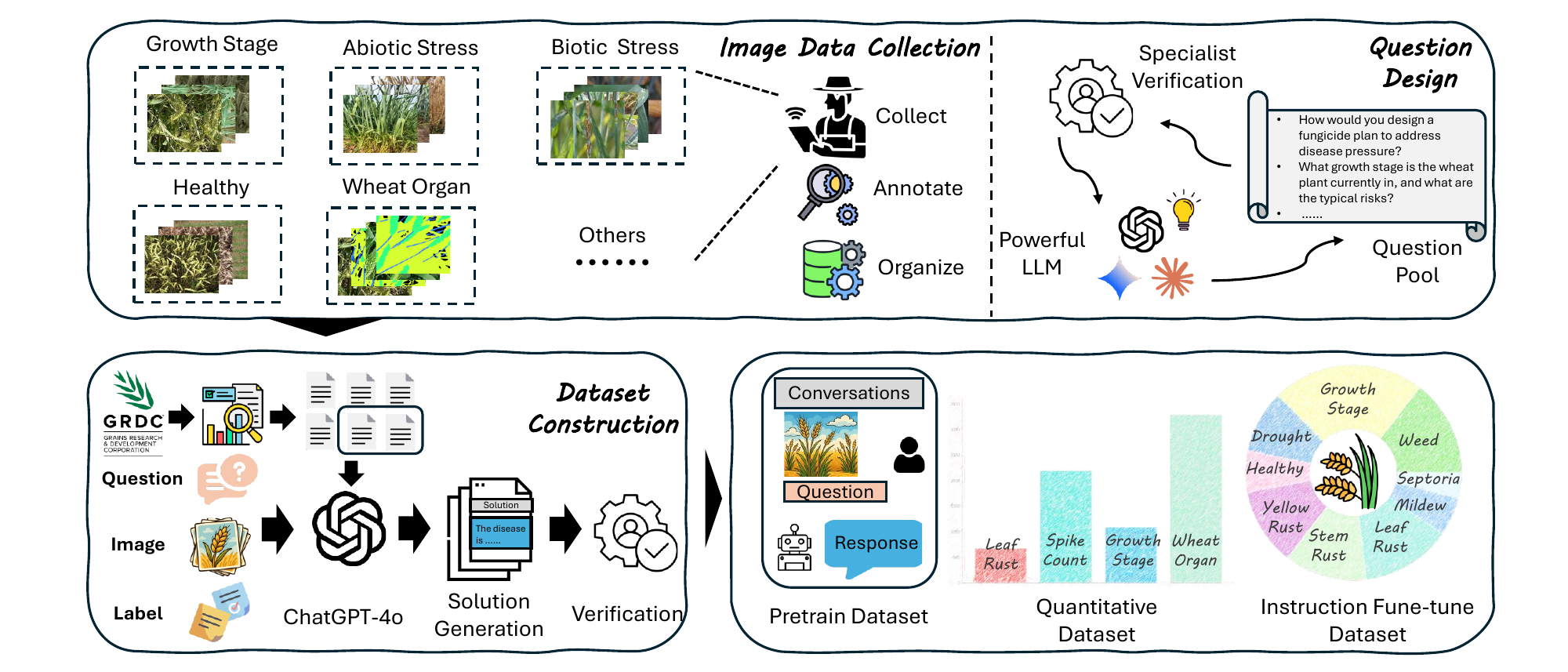}\vspace{-2ex}
    \caption{Overview of our agricultural visual reasoning dataset pipeline. \textmd{We collect and annotate wheat field images across various categories. A set of high-quality questions is generated through collaboration between domain experts and large language models. Using authoritative references and curated images, ChatGPT-4o produces expert-level responses, which are then verified. The resulting dataset includes: (1) a pretraining set with image–question–answer dialogues, (2) a quantitative dataset with structured outputs, and (3) an instruction fine-tuning set covering diverse wheat-specific agronomic tasks.}}
\label{fig:flowchart}
\end{figure*}

Traditionally, designing management plans often relies on expert visual inspection, which is often labour-intensive, susceptible to subjectivity, time-consuming, and difficult to scale. These limitations motivate researchers to turn to data-driven and computer vision technologies, aiming for scalable, precise, and rapid assessments, ultimately transforming field management strategies. Among recent technical innovations, Vision-Language Models (VLMs)~\cite{clip, blip, pali, llava, qwen_2_5_vl} are a burgeoning solution to provide real-time management recommendations based on field image input. These models are typically pretrained on large-scale image–text pairs to learn cross-modal correspondences, and subsequently refined through supervised fine-tuning and reinforcement learning to enhance reasoning and align with human preferences. While VLMs demonstrate strong performance on general tasks, their direct application to challenging domain-specific tasks often results in poor quantification capability and severe hallucinations, attributed to the distribution shift and lack of domain-specific knowledge. Moreover, the scarcity of high-quality, wheat-specific data remains a critical bottleneck in developing robust and well-adapted VLMs tailored to wheat-related applications.

In response, we propose WisWheat, a multimodal dataset designed to strengthen the quantification and reasoning capabilities of VLMs, ultimately enabling more reliable and actionable management recommendations. Our proposed WisWheat consists of three progressive, wheat-specific subsets: (i) a pretraining dataset composed of 47,871 multimodal instruction-following pairs that support coarse adaptation from general-purpose models to the wheat management domain, (ii) a quantification dataset of 7,263 aimed at empowering VLMs with core quantitative reasoning capability for trait measurement. (iii) a knowledge-infused instruction finetuning dataset consisting of 4,888 expert-curated image-question-response triplets to support contextual interpretation and agronomic reasoning for stress diagnosis and growth stage-specific management. These three datasets collectively prepare the VLMs for descriptive conversation, quantitative analysis, and agronomic decision recommendation. The contribution is summarised as follows:

\begin{itemize}[leftmargin=*]
    \item \textbf{A Tailored Multimodal Dataset} comprises over 60,022 wheat-specific image-text data across three progressive tiers to enhance VLM capabilities for wheat management applications. 
    \item \textbf{A Comprehensive Benchmark} is developed to assess a wide range of VLMs on wheat trait measurement and wheat management recommendations.
    \item \textbf{A Ready-to-Deploy Wheat-Specific VLM} that demonstrates improved quantification ability and knowledge expertise, enables effective management recommendations for wheat cultivation.
\end{itemize}

\section{Related Work}

\begin{table}[htbp]
  \centering
  \Large
\caption{Summary of data resources used. \textmd{Here, `IFT' stands for the Instruction Fine-Tuning Layer, `QL' is the Quantification Layer, and `PL' denotes the Pretraining Layer.}}
  \label{tab:resource_stat}
  \renewcommand{\arraystretch}{1.4}
  \resizebox{0.98\linewidth}{!}{
    \begin{tabular}{@{} l c c c c @{}}
      \toprule
      \textbf{Data Resource} & \textbf{Type} & \textbf{Label Type} & \textbf{\# Images} & \textbf{Layer} \\
      \midrule
      WFD \cite{disease} & Wheat Disease & Image Level & 2,414 & IFT \\
      Minimal Dataset \cite{drought} & Drought & Image Level & 450 & IFT \\
      GWHD 2021 \cite{gwhd} & Wheat Head & Object Level & 6,500 & QL \\
      GWFSS \cite{gwfss} & \makecell{Growth Stage \\ \& Segmentation} & Pixel Level & 1,096 & QL, IFT  \\
      Radish Weed Dataset \cite{radishwheat} & Weed & Object Level & 552 & IFT \\
      Cobbity Wheat \cite{cw1,cw2} & Weed & Object Level & 169 & IFT \\
      GWFSS Unlabeled \cite{gwfss} & General & Unlabeled & 47,871 & PL \\
      Rust Segmentation Dataset \cite{rustratio} & \makecell{Leaf Rust \\ Segmentation} & Pixel Level & 680 & QL \\
      \bottomrule
    \end{tabular}
  }
\end{table}


\textbf{Vision Language Models.} Vision language models (VLMs) are designed to integrate and process both textual and visual modalities. Early approaches \cite{clip, align} employed contrastive pretraining to align image-text representations. A major limitation arises that images and text feature alignment on global level makes fine-grained understanding challenging. Recent advances \cite{llava,blip,qwen_2_5_vl, paligemma} introduced model architectures that leverage cross-modal transformers to better integrate visual and textual information. In addition, with instruction tuning techniques \cite{instruct_blip, instructgpt} enable VLMs to follow specific visual instructions, generate structured responses, and better handle complex visual-textual tasks.


\begin{figure*}[!t]
\centering
    \includegraphics[width=\textwidth]{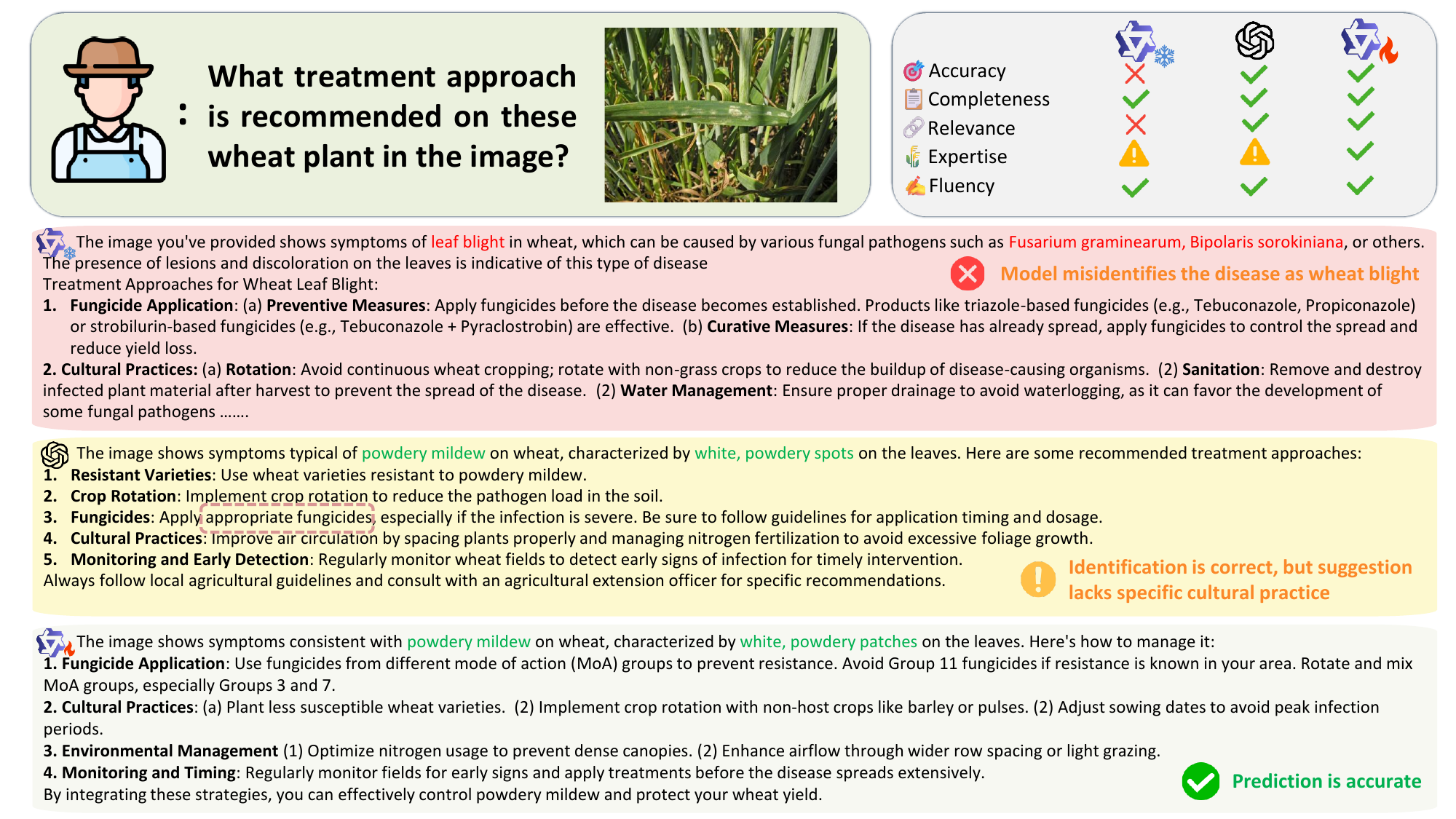}
    \caption{A case study on the model responses from Chatgpt-4o, base and fine-tuned Qwen2.5 VL 7B. \textmd{Base qwen2.5 VL 7B misidentifies the disease, and Chatgpt-4o response lacks specific cultural practice. On the other hand, Qwen2.5 VL 7B fine tuned on WisWheat provides accurate and complete analysis.}}
\label{fig:wheat_conversation_comparison}
\vspace{-0.6em}
\end{figure*}

\textbf{Domain-Specific Multimodal Datasets.} Domain-specific multimodal datasets have been developed to enhance vision-language model performance in specialized applications across various fields, including \cite{mimic_cxr, radio_vqa, vqa_medicine, pmc}, legal \cite{legalpile, lengalbench}, and agricultural domains \cite{agrieval,agri-llava, agrogpt}. Specifically in the agricultural domain, AgEval \cite{agrieval} focuses on VQA phenotyping tasks across identification, classification, and quantification, while Agri-LLaVA \cite{agri-llava} provides large scale plant diseases and pests dialogues for agricultural diagnostics. However, the existing datasets target general agricultural information rather than crop-specific management practices, or offer limited task coverage that lacks comprehensiveness for practical farming scenarios, which limits the specialized knowledge for specific crop species. Our work fills these gaps by introducing a wheat-specific multimodal dataset dedicated to wheat management practices, covering the complete wheat production cycle with detailed knowledge and management strategies.

\section{Dataset}
\label{sec:dataset}

\subsection{Dataset Overview}
To adapt VLMs for wheat management, we construct a multi-layered text-image dataset, spanning general domain alignment, task-specific quantification, and real-world management decision support. Our dataset, namely WisWheat, is organized into three distinct layers, each serving a complementary role in model training and evaluation. Specifically, we have: \\
\noindent\textbf{Pretraining Layer.} This fundamental layer comprises 47,871 image-text pairs designed to support foundational visual-semantic grounding. Each pair includes a wheat image and a descriptive-style prompt that encourages the extraction of visual attributes from the image. This dataset aims to establish coarse alignment between modalities within the general wheat field domain. By adapting VLM to diverse wheat field imagery and associated descriptions, this layer directly addresses the critical challenge of enabling the model to interpret and articulate input wheat-specific visual traits in natural language, providing essential grounding for downstream tasks.

\noindent\textbf{Quantification Layer.} The second layer focuses on empowering the VLMs with core quantitative reasoning capability in wheat phenotyping. In total, this layer consists of 7,263 image-question-answer triplets. In particular, this layer has 680 triplets for wheat rust coverage estimation, where responses are given as visual ratios; 2,200 triplets for spike counting, with responses provided in numeric form; 3,288 samples estimating wheat organ proportions, where answers are expressed as percentages; and 1,095 samples for growth stage classification, presented in multiple-choice format; For the growth stage classification, we follow the eight growth stage categorization as in ~\cite{gwfss}, including Emergence, Vegetative, Stem Elongation, Ear Emergence, Early Filling, Early Senescence, Late Senescence, and Maturity. We provide all eight stages in the growth stage question, and have one of them in the answer. Together, these tasks simulate real-world challenges in crop monitoring and support the development of models capable of translating complex phenotypic signals into interpretable, quantifiable outcomes across diverse field conditions.

\noindent\textbf{Instruction Fine-tuning Layer.} The last layer aims to facilitate wheat-management-focused conversations. This dataset includes 4,888 image-question-response triplets distributed across key topic categories. Each image is paired with a reasoning-style prompt and a response informed by curated knowledge sources, enabling the model to interpret subtle visual cues in a biologically meaningful way. For the biotic stressors, this dataset includes 692 samples for weed detection, 555 for yellow rust, 591 for stem rust, 644 for leaf rust, 277 for mildew, and 185 for septoria. Abiotic stress is represented by 450 drought-related samples, while 299 samples depict healthy wheat. Similar to the Quantification layer, the Instruction Fine-tuning layer contains 1,095 triplets that cover critical wheat growth stages. We note that, while these two layers share the common growth stage images, the question and response within each triplet are different. Growth stage triplets in this layer emphasize the management design choice, while in the Quantification layer, the triplet is for selecting the correct growth stage for any given image. By embedding real agronomic knowledge into the learning process, this layer ensures the model can generate accurate, context-aware insights and recommend actionable interventions, making it a valuable tool for supporting decision-making in field-based wheat phenotyping and crop management.

\subsection{Dataset Generation Process}
\textbf{Response Generation with Quality Control.} 
Fig.~\ref{fig:flowchart} illustrates the overall workflow for dataset construction. The process begins with the selection of image data from published datasets (as in Tab. \ref{tab:resource_stat}), accompanied by peer-reviewed research and covered under permissive public access licenses. An initial quality screening is then performed to remove images that are extremely low resolution, visually ambiguous or out of scope of our project focus. To ensure the quality of image-question-response triplets, we adopt an `expert in the loop' iterative generation approach. At each iteration, domain experts first assess the relevance and practical usefulness of the generated questions. Following this filtering step, responses for each retained image–question pair are generated using a combination of the original data source labels and authoritative references, such as official wheat fact sheets. To further improve response accuracy, we implement a confidence-based validation strategy: high-confidence correct or incorrect cases are automatically filtered and retained or discarded accordingly, while low-confidence or ambiguous triplets are flagged for expert review. This iterative process of prompt design, knowledge integration, specialist review, and dataset adaptation establishes these datasets as a robust, high-fidelity resource for multimodal instruction tuning in wheat stress diagnostics and crop management.

\noindent\textbf{Prompt Design} plays a critical role in adapting Vision-Language Models (VLMs) to wheat field management, guiding the model’s learning trajectory from general visual understanding to precise quantification and expert-level reasoning. For the pertaining layer, we developed a diverse set of general visual-language prompts, such as “Describe what you see in this image.” applied to images of wheat under various conditions. In contrast to the descriptive prompts used in pretraining, this Quantification Layer features specific, measurable questions. For instance, “How many wheat spikes are present?” or “What percentage of leaf area is affected by disease?” Each prompt was paired with image-level, object-level, or pixel-level annotation.
In the Instruction Fine-tuning Layer, the dataset focuses on contextual interpretation and actionable insight generation. Specifically, in this layer, we have questions like "What treatment approach is recommended on these wheat plant in the image?". For the full question list, we refer readers to our data repository.
\section{Modeling \& Training}

\subsection{Vision Language Model} 
VLMs integrate visual perception with language generation capabilities through an end-to-end architecture consisting of a visual encoder and a generative large language model (LLM). The visual encoder extracts a sequence of visual tokens $\mathbf{f}_v$ from the image input $\mathbf{X}_{Image}$. Based on the visual token, textual input, and previously generated tokens, the LLM predicts the next text token in an autoregressive manner. VLMs output until the end-of-sequence token (\texttt{EOF}) is produced. Formally, the generating process of VLM can be represented as:
\begin{equation}
\mathbf{X}_{out} = VLM(\mathbf{X}_{Image}, \mathbf{X}_{Text}),
\end{equation}
where $\mathbf{X}_{out}$ denotes the generated textual response. During inference, the model computes the conditional probability distribution of the next token $x_t$ over the vocabulary for each position $t$ in the sequence:

\begin{equation}
x_t \sim P(x_t|\mathbf{X}_{Image}, \mathbf{X}_{Text}, x_{<t}).
\end{equation}


\subsection{Domain-Specific Pretraining}
To enhance the model's wheat-specific representation learning and vision vision-language alignment, we pretrain VLMs using our wheat pretraining dataset containing large-scale wheat image-instruction pairs. However, direct fine-tune leads to catastrophic forgetting and overfitting problems. We follow \cite{llava} to leverage the projection layer for vision-language feature alignment, while keeping both the visual encoder and LLM frozen. In this way, wheat image features are aligned with LLM text embedding space within the wheat domain. This approach also preserves the pretrained capabilities for both vision encoder and language model, preparing model for subsequent instruction fine tuning on wheat management tasks.



\subsection{Chat Assistant Instruction Fine-tuning}
Following the domain-specific pretraining phase, we develop a specialized wheat assistant through instruction fine-tuning, using the curated wheat instruction dataset. We still freeze the vision encoder, while optimizing the feature alignment layer and the LLM. The fine-tuning dataset contains multi-round visual question-answer pairs and task-specific instructions related to topics including wheat assessment, stress identification and management, and growth stage analysis. We use cross-entropy loss on the model response tokens, formally expressed as
\begin{equation}
    \mathcal{L} = -\frac{1}{T} \sum_{t=1}^{T} \log P(x_t|\mathbf{X}_{Image}, \mathbf{X}_{Text}, x_{<t}),
\end{equation}
where T is the total sequence length. The fine-tuning process enables VLMs to transition from general wheat domain knowledge to specialized agricultural expertise. 


\begin{table*}[!htbp]
\centering
\Large
\caption{Wheat Conversation dataset benchmark results. \textmd{We evaluate open-sourced and closed-sourced base models, and SFT models on Wheat Conversation test dataset. The best results are highlighted in \textcolor{Maroon}{\textbf{Maroon bold}}, while the second-best results are highlighted in \textcolor{NavyBlue}{\underline{blue underlining}}.  Acc, Relev, Comp, Fluen, and Expert are the abbreviations for five evaluation criteria, \textit{i.e.}, Accuracy, Relevance, Completeness, Fluency, and Domain Expertise. $^\dag$ denotes that the model is fine-tuned with LoRA.}}
\label{tab:Chatbot}
\vspace{-0.7em}
\resizebox{0.88\linewidth}{!}{
\renewcommand{\arraystretch}{1.2}
\begin{tabular}{cc|cccccc|cccccc}
    \toprule
      & \textbf{Task} & \multicolumn{6}{c|}{\textbf{Wheat Stress}} & \multicolumn{6}{c}{\textbf{Growth Stage}} \\
    \midrule 
    & Model\textbackslash Criteria & \textbf{Acc.} & Relev. & Comp. & Fluen. & Expert. & \textbf{Avg.} & \textbf{Acc.} & Relev. & Comp. & Fluen. & Expert. & \textbf{Avg.} \\
    \midrule 
     \multirow{7}{*}{\textit{Open-eval}}
     & LLaVA 7B & 42.1 & 54.7 & 42.5 & 80.5 & 46.4 & 53.2 & 36.2 & 54.5 & 32.0 & 78.2 & 35.8 & 45.5  \\
     & LLaVA 13B & 45.6 & 57.8 & 45.2 & 82.3 & 49.3 & 56.0 & 33.6 & 45.5 & 33.6 & 78.4 & 36.0 & 45.4   \\
     & LLaVA Next Mistral 7B & 49.7 & 61.0 & 53.6 & 86.9 & 57.9 & 61.6 & 36.4 & 46.0 & 36.9 & 82.4 & 42.6 & 48.7  \\
     & LLaVA Next Vicuna 13B & 47.3 & 59.3 & 49.3 & 85.7 & 56.4 & 59.6 & 38.7 & 49.1 & 40.6 & 82.0 & 45.8 & 51.2  \\
     & Qwen2.5 VL 3B & 46.4 & 60.8 &	51.7 &	89.3 &	57.0 & 61.0 &	37.8 &	51.8 &	40.0 &	84.0 &	44.0 &	51.5 \\
     & Qwen2.5 VL 7B & 58.3 &	69.5 &	62.5 &	91.6 &	65.9 &	69.6 &	47.6 &	57.3 &	50.2 &	89.5 &	55.5 &	60.0  \\
     & Qwen2.5 VL 32B & 65.3 &	75.8 &	73.1 &	96.5 &	76.1 &	77.4 &	54.4 &	64.4 &	59.8 &	92.9 &	63.6 &	67.0  \\
    \midrule
     \multirow{3}{*}{\textit{Closed-eval}}
     & GPT-4o & 67.3 &	79.7 &	63.4 &	95.0 &	69.9 &	75.0 &	50.0 &	63.1 &	48.9 &	89.1 &	55.6 &	61.4 \\
     & Gemini 2.5 Pro & 72.5 &	82.0 &	76.3 &	96.8 &	83.6 &	82.2 &	67.5 &	78.4 &	72.6 &	96.2 &	80.2 &	79.0   \\
     & Claude Sonnect 3.7 & 60.5 & 69.5 &	60.3 &	90.6 &	66.0 &	69.4 &	46.4 &	54.4 &	45.6 &	86.0 &	50.9 &	56.7   \\
     \midrule 
     \multirow{5}{*}{\textit{Open-tuned}}
     & LLaVA 7B & \cellcolor{blue!10} 75.5 & \cellcolor{blue!10}	85.6 & \cellcolor{blue!10}	76.2 & \cellcolor{blue!10}	97.5 &	\cellcolor{blue!10} 83.8 & \cellcolor{blue!10}	83.7 & \cellcolor{blue!10}	80.0 & \cellcolor{blue!10}	89.3 & \cellcolor{blue!10}	78 & \cellcolor{blue!10} 97.5 & \cellcolor{blue!10}	85.3 & \cellcolor{blue!10}	86  \\
     & LLaVA 13B & \cellcolor{blue!10} \textcolor{NavyBlue}{\underline{76.8}} & \cellcolor{blue!10} \textcolor{NavyBlue}{\underline{86.5}} & \cellcolor{blue!10}	76.8 & \cellcolor{blue!10}	98.1 &	\cellcolor{blue!10} 84.4 & \cellcolor{blue!10}	\textcolor{NavyBlue}{\underline{84.5}} & \cellcolor{blue!10}	81.3 & \cellcolor{blue!10}	91.3 & \cellcolor{blue!10}	80.7 & \cellcolor{blue!10}	\textcolor{Maroon}{\textbf{99.3}} & \cellcolor{blue!10}	87.3 & \cellcolor{blue!10}	88.0 \\
     & Qwen2.5 VL 3B & \cellcolor{blue!10} 74.5 & \cellcolor{blue!10}	84.6 &	\cellcolor{blue!10} \textcolor{NavyBlue}{\underline{77.3}} & \cellcolor{blue!10}	98.5 & \cellcolor{blue!10}	84.8 & \cellcolor{blue!10}	83.9 & \cellcolor{blue!10}	\textcolor{NavyBlue}{\underline{82.7}} & \cellcolor{blue!10}	\textcolor{Maroon}{\textbf{92.4}} & \cellcolor{blue!10}	\textcolor{NavyBlue}{\underline{82.6}} &	\cellcolor{blue!10} 98.4 & \cellcolor{blue!10}	\textcolor{NavyBlue}{\underline{89.1}} & \cellcolor{blue!10}	\textcolor{NavyBlue}{\underline{89.0}} \\
     & Qwen2.5 VL 7B & \cellcolor{blue!10} \textcolor{Maroon}{\textbf{79.2}} & \cellcolor{blue!10}	\textcolor{Maroon}{\textbf{87.6}} & \cellcolor{blue!10}	\textcolor{Maroon}{\textbf{79.5}} & \cellcolor{blue!10}	\textcolor{Maroon}{\textbf{98.8}} & \cellcolor{blue!10}	\textcolor{Maroon}{\textbf{87.5}} & \cellcolor{blue!10}	\textcolor{Maroon}{\textbf{86.5}} & \cellcolor{blue!10}	\textcolor{Maroon}{\textbf{84.6}} & \cellcolor{blue!10}	\textcolor{NavyBlue}{\underline{91.6}} & \cellcolor{blue!10}	\textcolor{Maroon}{\textbf{82.7}} & \cellcolor{blue!10}	\textcolor{NavyBlue}{\underline{98.7}} & \cellcolor{blue!10}	\textcolor{Maroon}{\textbf{90}} & \cellcolor{blue!10}	\textcolor{Maroon}{\textbf{89.5}} \\
     & Qwen2.5 VL 32B$^\dag$ & \cellcolor{blue!10} 74.8 & \cellcolor{blue!10}	84.2 & \cellcolor{blue!10}	76.8 & \cellcolor{blue!10}	\textcolor{NavyBlue}{\underline{98.5}} & \cellcolor{blue!10}	\textcolor{NavyBlue}{\underline{85.2}} & \cellcolor{blue!10}	83.9 & \cellcolor{blue!10}	81.5 & \cellcolor{blue!10}	90.4 & \cellcolor{blue!10}	80.2 & \cellcolor{blue!10}	98.6 & \cellcolor{blue!10}	88.4 & \cellcolor{blue!10}	87.8  \\

    \bottomrule
\end{tabular}
}
\vspace{-0.7em}
\end{table*}

\subsection{Wheat VQA Reinforcement Learning}
Group Relative Policy Optimization (GRPO) \cite{deepseek_math} is a novel training approach on VLM visual question answering tasks \cite{visual_rft, vlmr1}. Unlike traditional reinforcement learning from human feedback (RLHF) methods that reply on human-annotated reward signals, GRPO uses automatically verifiable rewards that can be objectively determined. GRPO then compares multiple candidate responses to calculate relative advantages, simplifying reward and critic models.


We design distinct reward functions tailored to different task categories within the Wheat VQA dataset, which ensures appropriate optimization rewards for each task type. For the wheat growth stage classification task, where models should respond with exact answers, we implement a matching-based binary reward:

\begin{equation}
    R_{gs} = \begin{cases} 1, & \text{if } x_{\text{out}} = x_{\text{gt}} \\
    0, & \text{otherwise}
    \end{cases},
\end{equation}
where $x_{\text{out}}$ represents the VLM's extracted output answer from $\mathbf{X}_{out}$ and $x_{\text{gt}}$ denotes the ground truth.

For numerical prediction tasks including wheat spike counting, organ percentage estimation, and leaf rust severity assessment, we employ a continuous reward function that measures the relative proximity between predicted and ground truth values:

\begin{equation}
    R_{num} = e^{-\lambda\frac{|x_{\text{out}} - x_{\text{gt}}|}{x_{\text{gt}}}},
\end{equation}
where $\lambda$ is a scaling parameter.

In addition, we use a format reward across all tasks, which verifies if the responses adhere to the required structured format: \texttt{<think></think><answer>final answer</answer>}. The format reward facilitates both response consistency and reasoning ability.






   \section{Benchmark}
\subsection{Benchmark Setup}

We construct a diverse model pool that contains both closed-source and open-source Vision-Language Models (VLMs). Specifically, we incorporate three state-of-the-art closed-source commercial models in the model pool, including GPT-4o \cite{chatgpt_4o}, Claude 3.7 Sonnet \cite{claude2024}, and Gemini 2.5 Pro \cite{gemini}. These closed-source models will be directly evaluated on the test split of WisWheat through official APIs. We name this set of experiments as \textit{Closed-eval.} Our model pool also contains seven open-source VLMs, including the Qwen2.5 VL series \cite{qwen_2_5_vl} (3B, 7B, and 32B parameter variants), LLaVA 1.5 series \cite{llava} (7B and 13B), and LLaVA NEXT series \cite{liu2024llavanext} (LLaVA NEXT Mistral 7B and Vicuna 13B). We report two sets of performance for open-source models, namely \textit{Open-eval} and \textit{Open-tuned}. For the \textit{Open-eval}, the models are directly evaluated on the test split after loading the off-the-shelf weights from \href{https://huggingface.co}{Hugging Face}. For the \textit{Open-tuned}, we first train the model on the training split of our dataset, followed by the evaluation on the test split. The pretraining process of \textit{Open-tuned} runs for one epoch with a learning rate of 4e-5. In the fine-tuning phase, we maintain the vision encoder frozen while jointly training the adaptation layer and language decoder. We fine-tune using the wheat instruction dataset for three epochs with a learning rate of 1e-5. For particularly large models, such as Qwen2.5 VL 32B, we implement low-rank adaptation (LoRA) to efficiently fine-tune the models. For the VQA reinforcement learning process, we set the rollout size to 8 and the learning rate to 1e-6. All experiments are conducted on 4$\times$ NVIDIA H100 SXM 80GB GPUs.


\subsection{Wheat Multimodal Conversation}
\subsubsection{Evaluation Criteria}
To assess the performance of VLMs on the wheat multimodal instruction dataset, we leverage GPT-as-a-judge evaluation \cite{llm_as_judge, Q_align} that utilizes GPT-4o as an automated evaluator. For responses generated by VLMs, GPT-4o provides structured point-wise evaluation based on predefined criteria. Model response accuracy is primarily important for evaluating answer correctness and reliability. Beyond the core objective criterion, we incorporate additional criteria to reflect human preference, regarding the quality and utility from a user perspective. Overall, the score criteria evaluate VLM responses across five critical dimensions: \textbf{Accuracy:} Evaluates whether the answer provides correct and reliable information. \textbf{Completeness:} Measures how thoroughly the answer covers all aspects of the user's question. \textbf{Relevance:} Assesses how directly the answer addresses the user's question within the wheat field context and whether it maintains logical flow and coherence. \textbf{Domain Expertise:} Determines the level of specialized wheat research knowledge demonstrated, including appropriate terminology, methodologies, and contextual understanding of agricultural concepts specific to wheat. \textbf{Fluency:} Evaluates language quality through the clarity, naturalness, and readability of the writing.


As discussed in Sec.~\ref{sec:dataset}, the reference solutions are derived from expert wheat-specific knowledge curated with GPT-4o assistantce. During the evaluation process, we provide GPT-4o with reference solutions alongside detailed score criteria for test samples. For each dimension, GPT-4o assigns a score ranging from 1 to 5, where each level relates to a specific quality judgement.

\vspace{-1pt}
\subsubsection{Benchmark Results}
Tab.~\ref{tab:Chatbot} presents the performance comparison among \textit{Open-eval}, \textit{Closed-eval}, and \textit{Open-tuned}. The results demonstrate that \textit{Open-tuned} consistently outperforms the other two sets. In particular, Qwen2.5 VL 7B fine-tuned on our wheat instruction dataset achieves accuracy scores of 79.2\% and 84.6\% on wheat stress and growth stage conversation tasks, which significantly exceed the performance of ChatGPT-4o by margins of 11.9\% and 34.6\%. These results underscore the efficacy of the proposed dataset in enhancing model performance on wheat management-related conversational tasks. Fig.~\ref{fig:wheat_conversation_comparison} illustrates comparative responses from the base Qwen2.5 VL 7B model, fine-tuned Qwen2.5 VL 7B model, and ChatGPT-4o, on the same wheat image and question. When prompted to wheat diseases management, the base Qwen2.5 VL 7B incorrectly classifies powder mildew as leaf blight. On the other hand, while both our fine-tuned Qwen2.5 VL 7B and ChatGPT-4o correctly identify the leaf rust disease, their responses differ in practical utility. ChatGPT-4o provides generic management recommendations that lack specificity for the identified condition. In contrast, the fine-tuned model provides more comprehensive justification for its diagnosis along with targeted management strategies for mildew, demonstrating superior practical value.

\begin{table}[!t]
\centering
\Large
\caption{Wheat VQA dataset benchmark results. \textmd{We employ the GRPO to fine tune Qwen2.5 VL on Wheat VQA datasets, and we can observe significant performance gain from RL fine tuning, compared against base models and closed-source LLMs.}}
\label{tab:VQA}
\vspace{-0.6em}
\resizebox{0.98\linewidth}{!}{
\renewcommand{\arraystretch}{1.2}
\begin{tabular}{cc|cccc}
    \toprule
      & Model & \makecell{Growth\\Stage \textcolor{Green}{$\downarrow$}} & \makecell{Spike\\Count \textcolor{Green}{$\downarrow$}} & \makecell{Organ\\Ratio \textcolor{Green}{$\downarrow$}} & \makecell{Leaf Rust\\Ratio \textcolor{Green}{$\downarrow$}} \\
    \midrule 
     \multirow{3}{*}{\textit{Open-eval}}
     & Qwen2.5 VL 3B & 2.4 & 45.0 & 36.0 &  24.3 \\
     & Qwen2.5 VL 7B &  1.8 & 33.1 & 35.9 & 19.5  \\
     & Qwen2.5 VL 32B & 1.5 & 42.2 & 34.8 & 15.9  \\
    \midrule
     \multirow{3}{*}{\textit{Closed-eval}}
     & GPT-4o & 1.4  & 38.3 & 24.2 & \textcolor{NavyBlue}{\underline{7.2}} \\
     & Gemini 2.5 Pro &  1.0  & 30.4 & 21.9 & 11.6 \\
     & Claude Sonnect 3.7 & 1.5 & 36.8 &  26.2 & 11.3 \\
     \midrule 
     \multirow{2}{*}{\textit{Open-tuned}}
     &  Qwen2.5 VL 3B & \cellcolor{blue!10}\textcolor{NavyBlue}{\underline{0.9}}  & \cellcolor{blue!10}\textcolor{NavyBlue}{\underline{25.5}} & \cellcolor{blue!10}\textcolor{NavyBlue}{\underline{6.8}} &  \cellcolor{blue!10}{8.8} \\
     &  Qwen2.5 VL 7B & \cellcolor{blue!10}\textcolor{Maroon}{\textbf{0.7}}& \cellcolor{blue!10}\textcolor{Maroon}{\textbf{15.9}}& \cellcolor{blue!10}\textcolor{Maroon}{\textbf{5.5}}& \cellcolor{blue!10}\textcolor{Maroon}{\textbf{2.9}} \\

    \bottomrule
\end{tabular}
}
\vspace{-0.6em}
\end{table}

\subsection{Wheat Multimodal Quantification}
\subsubsection{Evaluation Criteria}
To assess model performance across diverse wheat VQA tasks, we employ task-wise evaluation metrics. For the wheat growth stage classification multiple-choice task, we employ Mean Absolute Error (MAE) instead of accuracy metrics. MAE leverages the inherently sequential progression of growth stages, where the numerical distance between predicted and actual stages provides meaningful information about prediction quality, such that predictions closer to true growth stages demonstrate better understanding compared to misclassification. For numerical prediction tasks including wheat spike counting, organ percentage estimation, and leaf rust severity assessment, we similarly employ MAE as the evaluation metric.


\subsubsection{Benchmark Results}
Tab.~\ref{tab:VQA} shows the experimental results of \textit{Open-eval}, \textit{Open-tuned}, and \textit{Closed-eval} across all Wheat VQA tasks. It can be observed that, without training on the proposed dataset, the base models exhibit limited capabilities across all tasks. For instance, the base Qwen2.5 VL 3B achieves 2.4 MAE on growth stage classification and 45.0 MAE on spike counting tasks. Closed-source models such as GPT-4o demonstrate marginal improvements over their respective base models, achieving a MAE of 1.4 in growth stage classification and 7.2 in rust percentage estimation. However, they still exhibit significant performance limitations without domain-specific fine-tuning for agricultural applications. This underscores the importance of domain-specific training for agricultural applications. In contrast, the models trained on the WisWheat consistently achieve superior performance across all VQA tasks. Particularly, the trained Qwen 2.5 VL 3B demonstrates considerable enhancement over the base model: growth stage classification MAE decreases from 2.4 to 0.9, while wheat organ percentage estimation improves from 36.0 to 6.8 MAE. The substantial performance gains demonstrate the effectiveness of the wheat VQA dataset in enhancing model quantification abilities for wheat assessment tasks.




\section{Conclusion} 
The paper introduces WisWheat, a comprehensive three-tier vision-language dataset crafted for wheat management applications. The dataset comprises over 60,022 curated data pairs that address the critical gap in domain-specific agricultural knowledge for VLMs. Through extensive benchmarking, we demonstrate that the existing models can exhibit limitations when applied to wheat management tasks, while models enhanced on WisWheat show significant improvement. Our work establishes a foundation for advancing AI in wheat management, and provides a valuable resource for future research in agricultural vision-language modelling. While our dataset provides comprehensive coverage of wheat management scenarios, the domain knowledge is primarily based on the Australian wheat production system. Different geographical regions employ distinct genotypes and management strategies to various conditions. Extending WisWheat to encompass more diverse wheat production systems remains a future direction.


\bibliographystyle{ACM-Reference-Format}
\bibliography{main}


\begin{thebibliography}{39}


\ifx \showCODEN    \undefined \def \showCODEN     #1{\unskip}     \fi
\ifx \showISBNx    \undefined \def \showISBNx     #1{\unskip}     \fi
\ifx \showISBNxiii \undefined \def \showISBNxiii  #1{\unskip}     \fi
\ifx \showISSN     \undefined \def \showISSN      #1{\unskip}     \fi
\ifx \showLCCN     \undefined \def \showLCCN      #1{\unskip}     \fi
\ifx \shownote     \undefined \def \shownote      #1{#1}          \fi
\ifx \showarticletitle \undefined \def \showarticletitle #1{#1}   \fi
\ifx \showURL      \undefined \def \showURL       {\relax}        \fi
\providecommand\bibfield[2]{#2}
\providecommand\bibinfo[2]{#2}
\providecommand\natexlab[1]{#1}
\providecommand\showeprint[2][]{arXiv:#2}

\bibitem[Anil et~al\mbox{.}(2023)]%
        {gemini}
\bibfield{author}{\bibinfo{person}{Rohan Anil}, \bibinfo{person}{Sebastian Borgeaud}, \bibinfo{person}{Yonghui Wu}, \bibinfo{person}{Jean{-}Baptiste Alayrac}, \bibinfo{person}{Jiahui Yu}, \bibinfo{person}{Radu Soricut}, \bibinfo{person}{Johan Schalkwyk}, \bibinfo{person}{Andrew~M. Dai}, \bibinfo{person}{Anja Hauth}, \bibinfo{person}{Katie Millican}, \bibinfo{person}{David Silver}, \bibinfo{person}{Slav Petrov}, \bibinfo{person}{Melvin Johnson}, \bibinfo{person}{Ioannis Antonoglou}, \bibinfo{person}{Julian Schrittwieser}, \bibinfo{person}{Amelia Glaese}, \bibinfo{person}{Jilin Chen}, \bibinfo{person}{Emily Pitler}, \bibinfo{person}{Timothy~P. Lillicrap}, \bibinfo{person}{Angeliki Lazaridou}, \bibinfo{person}{Orhan Firat}, \bibinfo{person}{James Molloy}, \bibinfo{person}{Michael Isard}, \bibinfo{person}{Paul~Ronald Barham}, \bibinfo{person}{Tom Hennigan}, \bibinfo{person}{Benjamin Lee}, \bibinfo{person}{Fabio Viola}, \bibinfo{person}{Malcolm Reynolds}, \bibinfo{person}{Yuanzhong Xu}, \bibinfo{person}{Ryan
  Doherty}, \bibinfo{person}{Eli Collins}, \bibinfo{person}{Clemens Meyer}, \bibinfo{person}{Eliza Rutherford}, \bibinfo{person}{Erica Moreira}, \bibinfo{person}{Kareem Ayoub}, \bibinfo{person}{Megha Goel}, \bibinfo{person}{George Tucker}, \bibinfo{person}{Enrique Piqueras}, \bibinfo{person}{Maxim Krikun}, \bibinfo{person}{Iain Barr}, \bibinfo{person}{Nikolay Savinov}, \bibinfo{person}{Ivo Danihelka}, \bibinfo{person}{Becca Roelofs}, \bibinfo{person}{Ana{\"{\i}}s White}, \bibinfo{person}{Anders Andreassen}, \bibinfo{person}{Tamara von Glehn}, \bibinfo{person}{Lakshman Yagati}, \bibinfo{person}{Mehran Kazemi}, \bibinfo{person}{Lucas Gonzalez}, \bibinfo{person}{Misha Khalman}, \bibinfo{person}{Jakub Sygnowski}, {and} \bibinfo{person}{et al.}} \bibinfo{year}{2023}\natexlab{}.
\newblock \showarticletitle{Gemini: {A} Family of Highly Capable Multimodal Models}.
\newblock \bibinfo{journal}{\emph{CoRR}} (\bibinfo{year}{2023}).
\newblock
\href{https://doi.org/10.48550/ARXIV.2312.11805}{doi:\nolinkurl{10.48550/ARXIV.2312.11805}}


\bibitem[Anthropic(2024)]%
        {claude2024}
\bibfield{author}{\bibinfo{person}{Anthropic}.} \bibinfo{year}{2024}\natexlab{}.
\newblock \bibinfo{title}{Claude}.
\newblock
\urldef\tempurl%
\url{https://claude.ai}
\showURL{%
\tempurl}
\newblock
\shownote{AI Assistant}.


\bibitem[Arshad et~al\mbox{.}(2025)]%
        {agrieval}
\bibfield{author}{\bibinfo{person}{Muhammad~Arbab Arshad}, \bibinfo{person}{Talukder~Zaki Jubery}, \bibinfo{person}{Tirtho Roy}, \bibinfo{person}{Rim Nassiri}, \bibinfo{person}{Asheesh~K. Singh}, \bibinfo{person}{Arti Singh}, \bibinfo{person}{Chinmay Hegde}, \bibinfo{person}{Baskar Ganapathysubramanian}, \bibinfo{person}{Aditya Balu}, \bibinfo{person}{Adarsh Krishnamurthy}, {and} \bibinfo{person}{Soumik Sarkar}.} \bibinfo{year}{2025}\natexlab{}.
\newblock \showarticletitle{Leveraging Vision Language Models for Specialized Agricultural Tasks}. In \bibinfo{booktitle}{\emph{WACV}}. \bibinfo{publisher}{{IEEE}}.
\newblock


\bibitem[Awais et~al\mbox{.}(2025)]%
        {agrogpt}
\bibfield{author}{\bibinfo{person}{Muhammad Awais}, \bibinfo{person}{Ali Husain Salem~Abdulla Alharthi}, \bibinfo{person}{Amandeep Kumar}, \bibinfo{person}{Hisham Cholakkal}, {and} \bibinfo{person}{Rao~Muhammad Anwer}.} \bibinfo{year}{2025}\natexlab{}.
\newblock \showarticletitle{Agrogpt: Efficient agricultural vision-language model with expert tuning}. In \bibinfo{booktitle}{\emph{WACV}}. IEEE, \bibinfo{pages}{5687--5696}.
\newblock


\bibitem[Bai et~al\mbox{.}(2025)]%
        {qwen_2_5_vl}
\bibfield{author}{\bibinfo{person}{Shuai Bai}, \bibinfo{person}{Keqin Chen}, \bibinfo{person}{Xuejing Liu}, \bibinfo{person}{Jialin Wang}, \bibinfo{person}{Wenbin Ge}, \bibinfo{person}{Sibo Song}, \bibinfo{person}{Kai Dang}, \bibinfo{person}{Peng Wang}, \bibinfo{person}{Shijie Wang}, \bibinfo{person}{Jun Tang}, \bibinfo{person}{Humen Zhong}, \bibinfo{person}{Yuanzhi Zhu}, \bibinfo{person}{Ming{-}Hsuan Yang}, \bibinfo{person}{Zhaohai Li}, \bibinfo{person}{Jianqiang Wan}, \bibinfo{person}{Pengfei Wang}, \bibinfo{person}{Wei Ding}, \bibinfo{person}{Zheren Fu}, \bibinfo{person}{Yiheng Xu}, \bibinfo{person}{Jiabo Ye}, \bibinfo{person}{Xi Zhang}, \bibinfo{person}{Tianbao Xie}, \bibinfo{person}{Zesen Cheng}, \bibinfo{person}{Hang Zhang}, \bibinfo{person}{Zhibo Yang}, \bibinfo{person}{Haiyang Xu}, {and} \bibinfo{person}{Junyang Lin}.} \bibinfo{year}{2025}\natexlab{}.
\newblock \showarticletitle{Qwen2.5-VL Technical Report}.
\newblock \bibinfo{journal}{\emph{CoRR}} (\bibinfo{year}{2025}).
\newblock
\href{https://doi.org/10.48550/ARXIV.2502.13923}{doi:\nolinkurl{10.48550/ARXIV.2502.13923}}


\bibitem[Ben~Abacha et~al\mbox{.}(2019)]%
        {vqa_medicine}
\bibfield{author}{\bibinfo{person}{Asma Ben~Abacha}, \bibinfo{person}{Sadid~A Hasan}, \bibinfo{person}{Vivek~V Datla}, \bibinfo{person}{Dina Demner-Fushman}, {and} \bibinfo{person}{Henning M{\"u}ller}.} \bibinfo{year}{2019}\natexlab{}.
\newblock \showarticletitle{Vqa-med: Overview of the medical visual question answering task at imageclef 2019}. In \bibinfo{booktitle}{\emph{Proceedings of CLEF (Conference and Labs of the Evaluation Forum) 2019 Working Notes}}. 9-12 September 2019.
\newblock


\bibitem[Beyer et~al\mbox{.}(2024)]%
        {paligemma}
\bibfield{author}{\bibinfo{person}{Lucas Beyer}, \bibinfo{person}{Andreas Steiner}, \bibinfo{person}{André~Susano Pinto}, \bibinfo{person}{Alexander Kolesnikov}, \bibinfo{person}{Xiao Wang}, \bibinfo{person}{Daniel Salz}, \bibinfo{person}{Maxim Neumann}, \bibinfo{person}{Ibrahim Alabdulmohsin}, \bibinfo{person}{Michael Tschannen}, \bibinfo{person}{Emanuele Bugliarello}, \bibinfo{person}{Thomas Unterthiner}, \bibinfo{person}{Daniel Keysers}, \bibinfo{person}{Skanda Koppula}, \bibinfo{person}{Fangyu Liu}, \bibinfo{person}{Adam Grycner}, \bibinfo{person}{Alexey Gritsenko}, \bibinfo{person}{Neil Houlsby}, \bibinfo{person}{Manoj Kumar}, \bibinfo{person}{Keran Rong}, \bibinfo{person}{Julian Eisenschlos}, \bibinfo{person}{Rishabh Kabra}, \bibinfo{person}{Matthias Bauer}, \bibinfo{person}{Matko Bošnjak}, \bibinfo{person}{Xi Chen}, \bibinfo{person}{Matthias Minderer}, \bibinfo{person}{Paul Voigtlaender}, \bibinfo{person}{Ioana Bica}, \bibinfo{person}{Ivana Balazevic}, \bibinfo{person}{Joan Puigcerver},
  \bibinfo{person}{Pinelopi Papalampidi}, \bibinfo{person}{Olivier Henaff}, \bibinfo{person}{Xi Xiong}, \bibinfo{person}{Radu Soricut}, \bibinfo{person}{Jeremiah Harmsen}, {and} \bibinfo{person}{Xiaohua Zhai}.} \bibinfo{year}{2024}\natexlab{}.
\newblock \bibinfo{title}{PaliGemma: A versatile 3B VLM for transfer}.
\newblock
\showeprint[arxiv]{2407.07726}
\urldef\tempurl%
\url{https://arxiv.org/abs/2407.07726}
\showURL{%
\tempurl}


\bibitem[Chen et~al\mbox{.}(2023)]%
        {pali}
\bibfield{author}{\bibinfo{person}{Xi Chen}, \bibinfo{person}{Xiao Wang}, \bibinfo{person}{Soravit Changpinyo}, \bibinfo{person}{A.~J. Piergiovanni}, \bibinfo{person}{Piotr Padlewski}, \bibinfo{person}{Daniel Salz}, \bibinfo{person}{Sebastian Goodman}, \bibinfo{person}{Adam Grycner}, \bibinfo{person}{Basil Mustafa}, \bibinfo{person}{Lucas Beyer}, \bibinfo{person}{Alexander Kolesnikov}, \bibinfo{person}{Joan Puigcerver}, \bibinfo{person}{Nan Ding}, \bibinfo{person}{Keran Rong}, \bibinfo{person}{Hassan Akbari}, \bibinfo{person}{Gaurav Mishra}, \bibinfo{person}{Linting Xue}, \bibinfo{person}{Ashish~V. Thapliyal}, \bibinfo{person}{James Bradbury}, {and} \bibinfo{person}{Weicheng Kuo}.} \bibinfo{year}{2023}\natexlab{}.
\newblock \showarticletitle{PaLI: {A} Jointly-Scaled Multilingual Language-Image Model}. In \bibinfo{booktitle}{\emph{ICLR}}.
\newblock


\bibitem[Coleman(2021a)]%
        {cw2}
\bibfield{author}{\bibinfo{person}{Guy Coleman}.} \bibinfo{year}{2021}\natexlab{a}.
\newblock \bibinfo{title}{{20200827 - Cobbity Wheat BFLY}}.
\newblock \bibinfo{howpublished}{\url{https://weed-ai.sydney.edu.au/datasets/3c363da3-6274-45e4-a0ce-b307cb0f89cc}}.
\newblock


\bibitem[Coleman(2021b)]%
        {cw1}
\bibfield{author}{\bibinfo{person}{Guy Coleman}.} \bibinfo{year}{2021}\natexlab{b}.
\newblock \bibinfo{title}{{20201014 - Cobbity Wheat BFLY}}.
\newblock \bibinfo{howpublished}{\url{https://weed-ai.sydney.edu.au/datasets/73468c19-b098-406a-86fa-df172caaec16}}.
\newblock


\bibitem[Cooper et~al\mbox{.}(2022)]%
        {gxexm}
\bibfield{author}{\bibinfo{person}{Mark Cooper}, \bibinfo{person}{Carlos~D Messina}, \bibinfo{person}{Tom Tang}, \bibinfo{person}{Carla Gho}, \bibinfo{person}{Owen~M Powell}, \bibinfo{person}{Dean~W Podlich}, \bibinfo{person}{Frank Technow}, {and} \bibinfo{person}{Graeme~L Hammer}.} \bibinfo{year}{2022}\natexlab{}.
\newblock \showarticletitle{Predicting Genotype$\times$ Environment$\times$ Management (G$\times$ E$\times$ M) interactions for the design of crop improvement strategies: integrating breeder, agronomist, and farmer perspectives}.
\newblock \bibinfo{journal}{\emph{Plant breeding reviews}}  \bibinfo{volume}{46} (\bibinfo{year}{2022}), \bibinfo{pages}{467--585}.
\newblock


\bibitem[Dai et~al\mbox{.}(2023)]%
        {instruct_blip}
\bibfield{author}{\bibinfo{person}{Wenliang Dai}, \bibinfo{person}{Junnan Li}, \bibinfo{person}{Dongxu Li}, \bibinfo{person}{Anthony Meng~Huat Tiong}, \bibinfo{person}{Junqi Zhao}, \bibinfo{person}{Weisheng Wang}, \bibinfo{person}{Boyang Li}, \bibinfo{person}{Pascale Fung}, {and} \bibinfo{person}{Steven Hoi}.} \bibinfo{year}{2023}\natexlab{}.
\newblock \bibinfo{title}{InstructBLIP: Towards General-purpose Vision-Language Models with Instruction Tuning}.
\newblock
\showeprint[arxiv]{2305.06500}
\urldef\tempurl%
\url{https://arxiv.org/abs/2305.06500}
\showURL{%
\tempurl}


\bibitem[David et~al\mbox{.}(2021)]%
        {gwhd}
\bibfield{author}{\bibinfo{person}{Etienne David}, \bibinfo{person}{Mario Serouart}, \bibinfo{person}{Daniel Smith}, \bibinfo{person}{Simon Madec}, \bibinfo{person}{Kaaviya Velumani}, \bibinfo{person}{Shouyang Liu}, \bibinfo{person}{Xu Wang}, \bibinfo{person}{Francisco Pinto}, \bibinfo{person}{Shahameh Shafiee}, \bibinfo{person}{Izzat~S.A. Tahir}, \bibinfo{person}{Hisashi Tsujimoto}, \bibinfo{person}{Shuhei Nasuda}, \bibinfo{person}{Bangyou Zheng}, \bibinfo{person}{Norbert Kirchgessner}, \bibinfo{person}{Helge Aasen}, \bibinfo{person}{Andreas Hund}, \bibinfo{person}{Pouria Sadhegi-Tehran}, \bibinfo{person}{Koichi Nagasawa}, \bibinfo{person}{Goro Ishikawa}, \bibinfo{person}{S{\'e}bastien Dandrifosse}, \bibinfo{person}{Alexis Carlier}, \bibinfo{person}{Benjamin Dumont}, \bibinfo{person}{Benoit Mercatoris}, \bibinfo{person}{Byron Evers}, \bibinfo{person}{Ken Kuroki}, \bibinfo{person}{Haozhou Wang}, \bibinfo{person}{Masanori Ishii}, {and} \bibinfo{person}{Wei Guo}.} \bibinfo{year}{2021}\natexlab{}.
\newblock \showarticletitle{Global Wheat Head Detection 2021: An Improved Dataset for Benchmarking Wheat Head Detection Methods}.
\newblock \bibinfo{journal}{\emph{Plant Phenomics}}  \bibinfo{volume}{2021} (\bibinfo{year}{2021}), \bibinfo{pages}{9846158}.
\newblock
\href{https://doi.org/10.34133/2021/9846158}{doi:\nolinkurl{10.34133/2021/9846158}}


\bibitem[Elias et~al\mbox{.}(2016)]%
        {gxe2}
\bibfield{author}{\bibinfo{person}{Ani~A Elias}, \bibinfo{person}{Kelly~R Robbins}, \bibinfo{person}{RW Doerge}, {and} \bibinfo{person}{Mitchell~R Tuinstra}.} \bibinfo{year}{2016}\natexlab{}.
\newblock \showarticletitle{Half a century of studying genotype$\times$ environment interactions in plant breeding experiments}.
\newblock \bibinfo{journal}{\emph{Crop Science}} \bibinfo{volume}{56}, \bibinfo{number}{5} (\bibinfo{year}{2016}), \bibinfo{pages}{2090--2105}.
\newblock


\bibitem[Erenstein et~al\mbox{.}(2022)]%
        {erenstein2022global}
\bibfield{author}{\bibinfo{person}{Olaf Erenstein}, \bibinfo{person}{Moti Jaleta}, \bibinfo{person}{Khondoker~Abdul Mottaleb}, \bibinfo{person}{Kai Sonder}, \bibinfo{person}{Jason Donovan}, {and} \bibinfo{person}{Hans-Joachim Braun}.} \bibinfo{year}{2022}\natexlab{}.
\newblock \showarticletitle{Global trends in wheat production, consumption and trade}.
\newblock In \bibinfo{booktitle}{\emph{Wheat improvement: food security in a changing climate}}. \bibinfo{publisher}{Springer International Publishing Cham}, \bibinfo{pages}{47--66}.
\newblock


\bibitem[Genaev et~al\mbox{.}(2021)]%
        {disease}
\bibfield{author}{\bibinfo{person}{Mikhail Genaev}, \bibinfo{person}{Ekaterina Skolotneva}, \bibinfo{person}{Ekaterina Gultyaeva}, \bibinfo{person}{Elena Orlova}, \bibinfo{person}{Nicolas Bechtold}, {and} \bibinfo{person}{Dmitry Afonnikov}.} \bibinfo{year}{2021}\natexlab{}.
\newblock \showarticletitle{Image-Based Wheat Fungi Diseases Identification by Deep Learning}.
\newblock \bibinfo{journal}{\emph{Plants}} \bibinfo{volume}{10}, \bibinfo{number}{8} (\bibinfo{year}{2021}), \bibinfo{pages}{1500}.
\newblock
\href{https://doi.org/10.3390/plants10081500}{doi:\nolinkurl{10.3390/plants10081500}}


\bibitem[Guha et~al\mbox{.}(2023)]%
        {lengalbench}
\bibfield{author}{\bibinfo{person}{Neel Guha}, \bibinfo{person}{Julian Nyarko}, \bibinfo{person}{Daniel~E. Ho}, \bibinfo{person}{Christopher Ré}, \bibinfo{person}{Adam Chilton}, \bibinfo{person}{Aditya Narayana}, \bibinfo{person}{Alex Chohlas-Wood}, \bibinfo{person}{Austin Peters}, \bibinfo{person}{Brandon Waldon}, \bibinfo{person}{Daniel~N. Rockmore}, \bibinfo{person}{Diego Zambrano}, \bibinfo{person}{Dmitry Talisman}, \bibinfo{person}{Enam Hoque}, \bibinfo{person}{Faiz Surani}, \bibinfo{person}{Frank Fagan}, \bibinfo{person}{Galit Sarfaty}, \bibinfo{person}{Gregory~M. Dickinson}, \bibinfo{person}{Haggai Porat}, \bibinfo{person}{Jason Hegland}, \bibinfo{person}{Jessica Wu}, \bibinfo{person}{Joe Nudell}, \bibinfo{person}{Joel Niklaus}, \bibinfo{person}{John Nay}, \bibinfo{person}{Jonathan~H. Choi}, \bibinfo{person}{Kevin Tobia}, \bibinfo{person}{Margaret Hagan}, \bibinfo{person}{Megan Ma}, \bibinfo{person}{Michael Livermore}, \bibinfo{person}{Nikon Rasumov-Rahe}, \bibinfo{person}{Nils Holzenberger},
  \bibinfo{person}{Noam Kolt}, \bibinfo{person}{Peter Henderson}, \bibinfo{person}{Sean Rehaag}, \bibinfo{person}{Sharad Goel}, \bibinfo{person}{Shang Gao}, \bibinfo{person}{Spencer Williams}, \bibinfo{person}{Sunny Gandhi}, \bibinfo{person}{Tom Zur}, \bibinfo{person}{Varun Iyer}, {and} \bibinfo{person}{Zehua Li}.} \bibinfo{year}{2023}\natexlab{}.
\newblock \bibinfo{title}{LegalBench: A Collaboratively Built Benchmark for Measuring Legal Reasoning in Large Language Models}.
\newblock
\showeprint[arxiv]{2308.11462}~[cs.CL]
\urldef\tempurl%
\url{https://arxiv.org/abs/2308.11462}
\showURL{%
\tempurl}


\bibitem[Hurst et~al\mbox{.}(2024)]%
        {chatgpt_4o}
\bibfield{author}{\bibinfo{person}{Aaron Hurst}, \bibinfo{person}{Adam Lerer}, \bibinfo{person}{Adam~P. Goucher}, \bibinfo{person}{Adam Perelman}, \bibinfo{person}{Aditya Ramesh}, \bibinfo{person}{Aidan Clark}, \bibinfo{person}{AJ Ostrow}, \bibinfo{person}{Akila Welihinda}, \bibinfo{person}{Alan Hayes}, \bibinfo{person}{Alec Radford}, \bibinfo{person}{Aleksander Madry}, \bibinfo{person}{Alex Baker{-}Whitcomb}, \bibinfo{person}{Alex Beutel}, \bibinfo{person}{Alex Borzunov}, \bibinfo{person}{Alex Carney}, \bibinfo{person}{Alex Chow}, \bibinfo{person}{Alex Kirillov}, \bibinfo{person}{Alex Nichol}, \bibinfo{person}{Alex Paino}, \bibinfo{person}{Alex Renzin}, \bibinfo{person}{Alex~Tachard Passos}, \bibinfo{person}{Alexander Kirillov}, \bibinfo{person}{Alexi Christakis}, \bibinfo{person}{Alexis Conneau}, \bibinfo{person}{Ali Kamali}, \bibinfo{person}{Allan Jabri}, \bibinfo{person}{Allison Moyer}, \bibinfo{person}{Allison Tam}, \bibinfo{person}{Amadou Crookes}, \bibinfo{person}{Amin Tootoonchian},
  \bibinfo{person}{Ananya Kumar}, \bibinfo{person}{Andrea Vallone}, \bibinfo{person}{Andrej Karpathy}, \bibinfo{person}{Andrew Braunstein}, \bibinfo{person}{Andrew Cann}, \bibinfo{person}{Andrew Codispoti}, \bibinfo{person}{Andrew Galu}, \bibinfo{person}{Andrew Kondrich}, \bibinfo{person}{Andrew Tulloch}, \bibinfo{person}{Andrey Mishchenko}, \bibinfo{person}{Angela Baek}, \bibinfo{person}{Angela Jiang}, \bibinfo{person}{Antoine Pelisse}, \bibinfo{person}{Antonia Woodford}, \bibinfo{person}{Anuj Gosalia}, \bibinfo{person}{Arka Dhar}, \bibinfo{person}{Ashley Pantuliano}, \bibinfo{person}{Avi Nayak}, \bibinfo{person}{Avital Oliver}, \bibinfo{person}{Barret Zoph}, \bibinfo{person}{Behrooz Ghorbani}, \bibinfo{person}{Ben Leimberger}, \bibinfo{person}{Ben Rossen}, \bibinfo{person}{Ben Sokolowsky}, \bibinfo{person}{Ben Wang}, \bibinfo{person}{Benjamin Zweig}, \bibinfo{person}{Beth Hoover}, \bibinfo{person}{Blake Samic}, \bibinfo{person}{Bob McGrew}, \bibinfo{person}{Bobby Spero}, \bibinfo{person}{Bogo Giertler},
  \bibinfo{person}{Bowen Cheng}, \bibinfo{person}{Brad Lightcap}, \bibinfo{person}{Brandon Walkin}, \bibinfo{person}{Brendan Quinn}, \bibinfo{person}{Brian Guarraci}, \bibinfo{person}{Brian Hsu}, \bibinfo{person}{Bright Kellogg}, \bibinfo{person}{Brydon Eastman}, \bibinfo{person}{Camillo Lugaresi}, \bibinfo{person}{Carroll~L. Wainwright}, \bibinfo{person}{Cary Bassin}, \bibinfo{person}{Cary Hudson}, \bibinfo{person}{Casey Chu}, \bibinfo{person}{Chad Nelson}, \bibinfo{person}{Chak Li}, \bibinfo{person}{Chan~Jun Shern}, \bibinfo{person}{Channing Conger}, \bibinfo{person}{Charlotte Barette}, \bibinfo{person}{Chelsea Voss}, \bibinfo{person}{Chen Ding}, \bibinfo{person}{Cheng Lu}, \bibinfo{person}{Chong Zhang}, \bibinfo{person}{Chris Beaumont}, \bibinfo{person}{Chris Hallacy}, \bibinfo{person}{Chris Koch}, \bibinfo{person}{Christian Gibson}, \bibinfo{person}{Christina Kim}, \bibinfo{person}{Christine Choi}, \bibinfo{person}{Christine McLeavey}, \bibinfo{person}{Christopher Hesse}, \bibinfo{person}{Claudia
  Fischer}, \bibinfo{person}{Clemens Winter}, \bibinfo{person}{Coley Czarnecki}, \bibinfo{person}{Colin Jarvis}, \bibinfo{person}{Colin Wei}, \bibinfo{person}{Constantin Koumouzelis}, {and} \bibinfo{person}{Dane Sherburn}.} \bibinfo{year}{2024}\natexlab{}.
\newblock \showarticletitle{GPT-4o System Card}.
\newblock \bibinfo{journal}{\emph{CoRR}} (\bibinfo{year}{2024}).
\newblock
\href{https://doi.org/10.48550/ARXIV.2410.21276}{doi:\nolinkurl{10.48550/ARXIV.2410.21276}}


\bibitem[Jia et~al\mbox{.}(2021)]%
        {align}
\bibfield{author}{\bibinfo{person}{Chao Jia}, \bibinfo{person}{Yinfei Yang}, \bibinfo{person}{Ye Xia}, \bibinfo{person}{Yi-Ting Chen}, \bibinfo{person}{Zarana Parekh}, \bibinfo{person}{Hieu Pham}, \bibinfo{person}{Quoc~V. Le}, \bibinfo{person}{Yunhsuan Sung}, \bibinfo{person}{Zhen Li}, {and} \bibinfo{person}{Tom Duerig}.} \bibinfo{year}{2021}\natexlab{}.
\newblock \bibinfo{title}{Scaling Up Visual and Vision-Language Representation Learning With Noisy Text Supervision}.
\newblock
\showeprint[arxiv]{2102.05918}
\urldef\tempurl%
\url{https://arxiv.org/abs/2102.05918}
\showURL{%
\tempurl}


\bibitem[Johnson et~al\mbox{.}(2019)]%
        {mimic_cxr}
\bibfield{author}{\bibinfo{person}{Alistair~EW Johnson}, \bibinfo{person}{Tom~J Pollard}, \bibinfo{person}{Seth~J Berkowitz}, \bibinfo{person}{Nathaniel~R Greenbaum}, \bibinfo{person}{Matthew~P Lungren}, \bibinfo{person}{Chih-ying Deng}, \bibinfo{person}{Roger~G Mark}, {and} \bibinfo{person}{Steven Horng}.} \bibinfo{year}{2019}\natexlab{}.
\newblock \showarticletitle{MIMIC-CXR, a de-identified publicly available database of chest radiographs with free-text reports}.
\newblock \bibinfo{journal}{\emph{Scientific data}} \bibinfo{volume}{6}, \bibinfo{number}{1} (\bibinfo{year}{2019}), \bibinfo{pages}{317}.
\newblock


\bibitem[Lau et~al\mbox{.}(2018)]%
        {radio_vqa}
\bibfield{author}{\bibinfo{person}{Jason~J Lau}, \bibinfo{person}{Soumya Gayen}, \bibinfo{person}{Asma Ben~Abacha}, {and} \bibinfo{person}{Dina Demner-Fushman}.} \bibinfo{year}{2018}\natexlab{}.
\newblock \showarticletitle{A dataset of clinically generated visual questions and answers about radiology images}.
\newblock \bibinfo{journal}{\emph{Scientific data}} \bibinfo{volume}{5}, \bibinfo{number}{1} (\bibinfo{year}{2018}), \bibinfo{pages}{1--10}.
\newblock


\bibitem[Li et~al\mbox{.}(2022)]%
        {blip}
\bibfield{author}{\bibinfo{person}{Junnan Li}, \bibinfo{person}{Dongxu Li}, \bibinfo{person}{Caiming Xiong}, {and} \bibinfo{person}{Steven C.~H. Hoi}.} \bibinfo{year}{2022}\natexlab{}.
\newblock \showarticletitle{{BLIP:} Bootstrapping Language-Image Pre-training for Unified Vision-Language Understanding and Generation}. In \bibinfo{booktitle}{\emph{ICML}}, Vol.~\bibinfo{volume}{162}. \bibinfo{publisher}{{PMLR}}, \bibinfo{pages}{12888--12900}.
\newblock


\bibitem[Liu et~al\mbox{.}(2024)]%
        {liu2024llavanext}
\bibfield{author}{\bibinfo{person}{Haotian Liu}, \bibinfo{person}{Chunyuan Li}, \bibinfo{person}{Yuheng Li}, \bibinfo{person}{Bo Li}, \bibinfo{person}{Yuanhan Zhang}, \bibinfo{person}{Sheng Shen}, {and} \bibinfo{person}{Yong~Jae Lee}.} \bibinfo{year}{2024}\natexlab{}.
\newblock \bibinfo{title}{LLaVA-NeXT: Improved reasoning, OCR, and world knowledge}.
\newblock
\urldef\tempurl%
\url{https://llava-vl.github.io/blog/2024-01-30-llava-next/}
\showURL{%
\tempurl}


\bibitem[Liu et~al\mbox{.}(2023)]%
        {llava}
\bibfield{author}{\bibinfo{person}{Haotian Liu}, \bibinfo{person}{Chunyuan Li}, \bibinfo{person}{Qingyang Wu}, {and} \bibinfo{person}{Yong~Jae Lee}.} \bibinfo{year}{2023}\natexlab{}.
\newblock \showarticletitle{Visual Instruction Tuning}. In \bibinfo{booktitle}{\emph{NeurIPS}}.
\newblock


\bibitem[Liu et~al\mbox{.}(2025)]%
        {visual_rft}
\bibfield{author}{\bibinfo{person}{Ziyu Liu}, \bibinfo{person}{Zeyi Sun}, \bibinfo{person}{Yuhang Zang}, \bibinfo{person}{Xiaoyi Dong}, \bibinfo{person}{Yuhang Cao}, \bibinfo{person}{Haodong Duan}, \bibinfo{person}{Dahua Lin}, {and} \bibinfo{person}{Jiaqi Wang}.} \bibinfo{year}{2025}\natexlab{}.
\newblock \showarticletitle{Visual-RFT: Visual Reinforcement Fine-Tuning}.
\newblock \bibinfo{journal}{\emph{CoRR}} (\bibinfo{year}{2025}).
\newblock
\href{https://doi.org/10.48550/ARXIV.2503.01785}{doi:\nolinkurl{10.48550/ARXIV.2503.01785}}


\bibitem[Niklaus et~al\mbox{.}(2023)]%
        {legalpile}
\bibfield{author}{\bibinfo{person}{Joel Niklaus}, \bibinfo{person}{Veton Matoshi}, \bibinfo{person}{Matthias St{\"u}rmer}, \bibinfo{person}{Ilias Chalkidis}, {and} \bibinfo{person}{Daniel~E Ho}.} \bibinfo{year}{2023}\natexlab{}.
\newblock \showarticletitle{Multilegalpile: A 689gb multilingual legal corpus}.
\newblock \bibinfo{journal}{\emph{arXiv preprint arXiv:2306.02069}} (\bibinfo{year}{2023}).
\newblock


\bibitem[Ouyang et~al\mbox{.}(2022)]%
        {instructgpt}
\bibfield{author}{\bibinfo{person}{Long Ouyang}, \bibinfo{person}{Jeff Wu}, \bibinfo{person}{Xu Jiang}, \bibinfo{person}{Diogo Almeida}, \bibinfo{person}{Carroll~L. Wainwright}, \bibinfo{person}{Pamela Mishkin}, \bibinfo{person}{Chong Zhang}, \bibinfo{person}{Sandhini Agarwal}, \bibinfo{person}{Katarina Slama}, \bibinfo{person}{Alex Ray}, \bibinfo{person}{John Schulman}, \bibinfo{person}{Jacob Hilton}, \bibinfo{person}{Fraser Kelton}, \bibinfo{person}{Luke Miller}, \bibinfo{person}{Maddie Simens}, \bibinfo{person}{Amanda Askell}, \bibinfo{person}{Peter Welinder}, \bibinfo{person}{Paul Christiano}, \bibinfo{person}{Jan Leike}, {and} \bibinfo{person}{Ryan Lowe}.} \bibinfo{year}{2022}\natexlab{}.
\newblock \bibinfo{title}{Training language models to follow instructions with human feedback}.
\newblock
\showeprint[arxiv]{2203.02155}
\urldef\tempurl%
\url{https://arxiv.org/abs/2203.02155}
\showURL{%
\tempurl}


\bibitem[Radford et~al\mbox{.}(2021)]%
        {clip}
\bibfield{author}{\bibinfo{person}{Alec Radford}, \bibinfo{person}{Jong~Wook Kim}, \bibinfo{person}{Chris Hallacy}, \bibinfo{person}{Aditya Ramesh}, \bibinfo{person}{Gabriel Goh}, \bibinfo{person}{Sandhini Agarwal}, \bibinfo{person}{Girish Sastry}, \bibinfo{person}{Amanda Askell}, \bibinfo{person}{Pamela Mishkin}, \bibinfo{person}{Jack Clark}, \bibinfo{person}{Gretchen Krueger}, {and} \bibinfo{person}{Ilya Sutskever}.} \bibinfo{year}{2021}\natexlab{}.
\newblock \showarticletitle{Learning Transferable Visual Models From Natural Language Supervision}, Vol.~\bibinfo{volume}{139}. \bibinfo{publisher}{{PMLR}}, \bibinfo{pages}{8748--8763}.
\newblock


\bibitem[Rayner(2022)]%
        {radishwheat}
\bibfield{author}{\bibinfo{person}{Gilbert Rayner}.} \bibinfo{year}{2022}\natexlab{}.
\newblock \bibinfo{title}{{RadishWheatDataset}}.
\newblock \bibinfo{howpublished}{\url{https://weed-ai.sydney.edu.au/datasets/8b8f134f-ede4-4792-b1f7-d38fc05d8127}}.
\newblock


\bibitem[Shao et~al\mbox{.}(2024)]%
        {deepseek_math}
\bibfield{author}{\bibinfo{person}{Zhihong Shao}, \bibinfo{person}{Peiyi Wang}, \bibinfo{person}{Qihao Zhu}, \bibinfo{person}{Runxin Xu}, \bibinfo{person}{Junxiao Song}, \bibinfo{person}{Mingchuan Zhang}, \bibinfo{person}{Y.~K. Li}, \bibinfo{person}{Y. Wu}, {and} \bibinfo{person}{Daya Guo}.} \bibinfo{year}{2024}\natexlab{}.
\newblock \showarticletitle{DeepSeekMath: Pushing the Limits of Mathematical Reasoning in Open Language Models}.
\newblock \bibinfo{journal}{\emph{CoRR}}  \bibinfo{volume}{abs/2402.03300} (\bibinfo{year}{2024}).
\newblock
\href{https://doi.org/10.48550/ARXIV.2402.03300}{doi:\nolinkurl{10.48550/ARXIV.2402.03300}}


\bibitem[Shen et~al\mbox{.}(2025)]%
        {vlmr1}
\bibfield{author}{\bibinfo{person}{Haozhan Shen}, \bibinfo{person}{Peng Liu}, \bibinfo{person}{Jingcheng Li}, \bibinfo{person}{Chunxin Fang}, \bibinfo{person}{Yibo Ma}, \bibinfo{person}{Jiajia Liao}, \bibinfo{person}{Qiaoli Shen}, \bibinfo{person}{Zilun Zhang}, \bibinfo{person}{Kangjia Zhao}, \bibinfo{person}{Qianqian Zhang}, \bibinfo{person}{Ruochen Xu}, {and} \bibinfo{person}{Tiancheng Zhao}.} \bibinfo{year}{2025}\natexlab{}.
\newblock \bibinfo{title}{VLM-R1: A Stable and Generalizable R1-style Large Vision-Language Model}.
\newblock


\bibitem[Wang et~al\mbox{.}(2024)]%
        {agri-llava}
\bibfield{author}{\bibinfo{person}{Liqiong Wang}, \bibinfo{person}{Teng Jin}, \bibinfo{person}{Jinyu Yang}, \bibinfo{person}{Ales Leonardis}, \bibinfo{person}{Fangyi Wang}, {and} \bibinfo{person}{Feng Zheng}.} \bibinfo{year}{2024}\natexlab{}.
\newblock \showarticletitle{Agri-LLaVA: Knowledge-Infused Large Multimodal Assistant on Agricultural Pests and Diseases}.
\newblock \bibinfo{journal}{\emph{CoRR}} (\bibinfo{year}{2024}).
\newblock


\bibitem[Wang et~al\mbox{.}(2025)]%
        {gwfss}
\bibfield{author}{\bibinfo{person}{Zijian Wang}, \bibinfo{person}{Radek Zenkl}, \bibinfo{person}{Latifa Greche}, \bibinfo{person}{Benoit De~Solan}, \bibinfo{person}{Lucas~Bernigaud Samatan}, \bibinfo{person}{Safaa Ouahid}, \bibinfo{person}{Andrea Visioni}, \bibinfo{person}{Carlos~A Robles-Zazueta}, \bibinfo{person}{Francisco Pinto}, \bibinfo{person}{Ivan Perez-Olivera}, {et~al\mbox{.}}} \bibinfo{year}{2025}\natexlab{}.
\newblock \showarticletitle{The Global Wheat Full Semantic Organ Segmentation (GWFSS) Dataset}.
\newblock \bibinfo{journal}{\emph{bioRxiv}} (\bibinfo{year}{2025}), \bibinfo{pages}{2025--03}.
\newblock


\bibitem[Williams et~al\mbox{.}(2008)]%
        {gxe1}
\bibfield{author}{\bibinfo{person}{RM Williams}, \bibinfo{person}{Louise O’Brien}, \bibinfo{person}{Howard~A Eagles}, \bibinfo{person}{Vicky~A Solah}, {and} \bibinfo{person}{Vijay Jayasena}.} \bibinfo{year}{2008}\natexlab{}.
\newblock \showarticletitle{The influences of genotype, environment, and genotype$\times$ environment interaction on wheat quality}.
\newblock \bibinfo{journal}{\emph{Australian journal of agricultural research}} \bibinfo{volume}{59}, \bibinfo{number}{2} (\bibinfo{year}{2008}), \bibinfo{pages}{95--111}.
\newblock


\bibitem[Wu et~al\mbox{.}(2023)]%
        {pmc}
\bibfield{author}{\bibinfo{person}{Chaoyi Wu}, \bibinfo{person}{Weixiong Lin}, \bibinfo{person}{Xiaoman Zhang}, \bibinfo{person}{Ya Zhang}, \bibinfo{person}{Yanfeng Wang}, {and} \bibinfo{person}{Weidi Xie}.} \bibinfo{year}{2023}\natexlab{}.
\newblock \bibinfo{title}{PMC-LLaMA: Towards Building Open-source Language Models for Medicine}.
\newblock
\showeprint[arxiv]{2304.14454}~[cs.CL]
\urldef\tempurl%
\url{https://arxiv.org/abs/2304.14454}
\showURL{%
\tempurl}


\bibitem[Wu et~al\mbox{.}(2024)]%
        {Q_align}
\bibfield{author}{\bibinfo{person}{Haoning Wu}, \bibinfo{person}{Zicheng Zhang}, \bibinfo{person}{Weixia Zhang}, \bibinfo{person}{Chaofeng Chen}, \bibinfo{person}{Liang Liao}, \bibinfo{person}{Chunyi Li}, \bibinfo{person}{Yixuan Gao}, \bibinfo{person}{Annan Wang}, \bibinfo{person}{Erli Zhang}, \bibinfo{person}{Wenxiu Sun}, \bibinfo{person}{Qiong Yan}, \bibinfo{person}{Xiongkuo Min}, \bibinfo{person}{Guangtao Zhai}, {and} \bibinfo{person}{Weisi Lin}.} \bibinfo{year}{2024}\natexlab{}.
\newblock \showarticletitle{Q-Align: Teaching LMMs for Visual Scoring via Discrete Text-Defined Levels}. In \bibinfo{booktitle}{\emph{ICML}}.
\newblock


\bibitem[Yao et~al\mbox{.}(2024)]%
        {drought}
\bibfield{author}{\bibinfo{person}{Jianbin Yao}, \bibinfo{person}{Yushu Wu}, \bibinfo{person}{Jianhua Liu}, {and} \bibinfo{person}{Hansheng Wang}.} \bibinfo{year}{2024}\natexlab{}.
\newblock \showarticletitle{Multimodal deep learning-based drought monitoring research for winter wheat during critical growth stages}.
\newblock \bibinfo{journal}{\emph{PLOS ONE}} \bibinfo{volume}{19}, \bibinfo{number}{5} (\bibinfo{year}{2024}), \bibinfo{pages}{e0300746}.
\newblock
\href{https://doi.org/10.1371/journal.pone.0300746}{doi:\nolinkurl{10.1371/journal.pone.0300746}}


\bibitem[Zenkl et~al\mbox{.}(2025)]%
        {rustratio}
\bibfield{author}{\bibinfo{person}{Radek Zenkl}, \bibinfo{person}{Bruce~A. McDonald}, \bibinfo{person}{Achim Walter}, {and} \bibinfo{person}{Jonas Anderegg}.} \bibinfo{year}{2025}\natexlab{}.
\newblock \showarticletitle{Towards high throughput in-field detection and quantification of wheat foliar diseases using deep learning}.
\newblock \bibinfo{journal}{\emph{Computers and Electronics in Agriculture}}  \bibinfo{volume}{232} (\bibinfo{year}{2025}), \bibinfo{pages}{109854}.
\newblock
\href{https://doi.org/10.1016/j.compag.2024.109854}{doi:\nolinkurl{10.1016/j.compag.2024.109854}}


\bibitem[Zheng et~al\mbox{.}(2023)]%
        {llm_as_judge}
\bibfield{author}{\bibinfo{person}{Lianmin Zheng}, \bibinfo{person}{Wei{-}Lin Chiang}, \bibinfo{person}{Ying Sheng}, \bibinfo{person}{Siyuan Zhuang}, \bibinfo{person}{Zhanghao Wu}, \bibinfo{person}{Yonghao Zhuang}, \bibinfo{person}{Zi Lin}, \bibinfo{person}{Zhuohan Li}, \bibinfo{person}{Dacheng Li}, \bibinfo{person}{Eric~P. Xing}, \bibinfo{person}{Hao Zhang}, \bibinfo{person}{Joseph~E. Gonzalez}, {and} \bibinfo{person}{Ion Stoica}.} \bibinfo{year}{2023}\natexlab{}.
\newblock \showarticletitle{Judging LLM-as-a-Judge with MT-Bench and Chatbot Arena}. In \bibinfo{booktitle}{\emph{NeurIPS}}.
\newblock


\end{thebibliography}










\end{document}